\documentclass[10pt,twocolumn,letterpaper]{article}

\usepackage{wacv}
\usepackage{times}
\usepackage{epsfig}
\usepackage{graphicx}
\usepackage{amsmath}
\usepackage{amssymb}

\usepackage{CJKutf8}
\usepackage{multirow}
\usepackage{comment}
\usepackage[caption=false]{subfig}
\usepackage{adjustbox}
\usepackage{blindtext}
\usepackage{enumitem}
\usepackage[dvipsnames]{xcolor}
\usepackage[pagebackref=true,breaklinks=true,letterpaper=true,colorlinks,bookmarks=false]{hyperref}

\wacvfinalcopy % *** Uncomment this line for the final submission

\def\wacvPaperID{330} % *** Enter the wacv Paper ID here

\ifwacvfinal\fi
\begin{document}
\begin{CJK}{UTF8}{mj}

%%%%%%%%% TITLE
\title{Combination of Multiple Global Descriptors for Image Retrieval}

% Authors at different institutions
\author{HeeJae Jun\thanks{Authors contributed equally}
\and
Byungsoo Ko\footnotemark[1]\\
\and
Youngjoon Kim\\
\and
Insik Kim\\
\and
Jongtack Kim\\
\and
NAVER/LINE Vision\\
{\tt\small \{heejae.jun, byungsoo.ko, kim.youngjoon, insik.kim, jongtack.kim\}@navercorp.com}
}

\maketitle
\ifwacvfinal\thispagestyle{empty}\fi

%%%%%%%%% ABSTRACT
\begin{abstract}
Recent studies in image retrieval task have shown that ensembling different models and combining multiple global descriptors lead to performance improvement.
However, training different models for the ensemble is not only difficult but also inefficient with respect to time and memory.
In this paper, we propose a novel framework that exploits multiple global descriptors to get an ensemble effect while it can be trained in an end-to-end manner.
The proposed framework is flexible and expandable by the global descriptor, CNN backbone, loss, and dataset.
Moreover, we investigate the effectiveness of combining multiple global descriptors with quantitative and qualitative analysis.
Our extensive experiments show that the combined descriptor outperforms a single global descriptor, as it can utilize different types of feature properties.
In the benchmark evaluation, the proposed framework achieves the state-of-the-art performance on the CARS196, CUB200-2011, In-shop Clothes, and Stanford Online Products on image retrieval tasks.
Our model implementations and pretrained models are publicly available\footnote{\url{https://github.com/naver/cgd}}.
% by a large margin compared to competing approaches.
\end{abstract}

%%%%%%%%% BODY TEXT
\section{Introduction}
% Image descriptor based on deep CNNs가 최근 다양한 분야에서 많이 쓰이고 있다.
Since the ground-breaking in 2012 ImageNet competition~\cite{deng2009imagenet, krizhevsky2012imagenet}, image descriptors based on deep convolutional neural networks (CNNs) have surfaced as generic descriptors in computer vision tasks, including classification~\cite{krizhevsky2012imagenet, he2016deep, szegedy2016rethinking}, object detection~\cite{girshick2015fast, ren2015faster, redmon2016you}, and semantic segmentation~\cite{long2015fully, chen2018deeplab, ronneberger2015u}.
% They have been adopted to highly semantic tasks as well, such as image captioning~\cite{you2016image, yao2017boosting} and visual question answering~\cite{antol2015vqa, lu2016hierarchical}.
Moreover, recent works leveraging image descriptors based on deep CNNs have emerged for image retrieval task which used to apply conventional methods relying on local descriptor matching~\cite{lowe2004distinctive, ke2004pca} and re-ranking with spatial verification~\cite{mikulik2010learning, tolias2016image, li2015pairwise}.

% Global descriptor
In the case of recent researches on image retrieval~\cite{babenko2014neural, gordo2016deep}, fully connected (FC) layers after several convolutional layers are used as global descriptors followed by dimensionality reduction.
Other works generate global descriptors from the activations of the convolutional layers.
Representative global descriptors generated by global-pooling methods include sum pooling of convolutions (SPoC)~\cite{Babenko_2015_ICCV}, maximum activation of convolutions (MAC)~\cite{tolias2015particular}, and generalized-mean pooling (GeM)~\cite{radenovic2018fine}.
The performance of each global descriptor varies by dataset as each descriptor has different properties~\cite{boureau2010theoretical}.
For example, SPoC activates larger regions on the image representation while MAC activates more focused regions~\cite{hoang2017selective}.
% In order to boost their ability, variants of these representative global descriptors have been proposed such as weighted sum pooling~\cite{kalantidis2016cross}, weighted GeM~\cite{wu2018weighted}, regional MAC (R-MAC)~\cite{tolias2015particular}, etc.

% 성능을 높이기 위해 ensemble하는 시도들이 있었고, 이러한 시도들에는 문제점이 있다. (ensemble, diversity 정의)
Recent researches have focused on ensemble techniques for image retrieval task.
Conventional ensemble techniques which train multiple learners individually and use a combined model lead to an increase in performance~\cite{lin2018regional, Xuan_2018_ECCV, opitz2017bier, Kim_2018_ECCV}.
Many high-ranked approaches~\cite{ozaki2019large, chen20192nd} in the recent Google landmark retrieval challenge~\cite{glc2019} and Zehang \textit{et al.}~\cite{lin2018regional} boost the performance by combining different global descriptors which are trained individually.
However, explicitly training multiple learners for ensemble could lead to longer training time and higher memory consumption.
In order to handle this problem, other ensemble approaches~\cite{Kim_2018_ECCV, opitz2017bier} attempt to train a retrieval model in an end-to-end manner.
These approaches can be tricky as they need specifically designed strategy or loss to control diversity among learners, which also cause a more laborious training process.

% ※ 추후 최종본 수정 때를 위한 참고용으로 기록
% Figure 1. 의 FC, BN, L2, FC+softmax 표기가 현재 표기 위치의 바로 앞 화살표에 붙어야 함
\begin{figure*}[h!t!]
\vspace{-1em}
\begin{center}
\includegraphics[width=1.\textwidth]{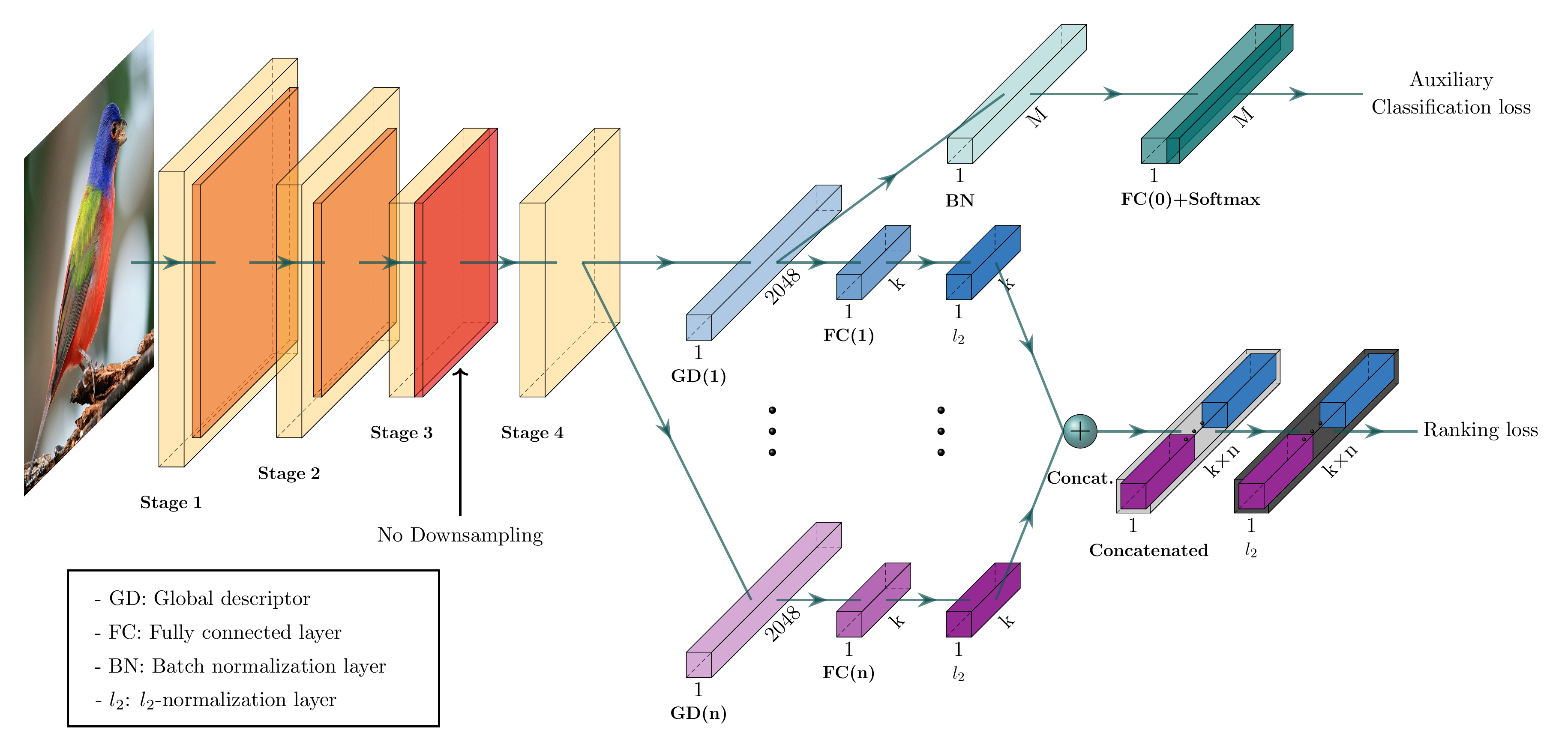}
\end{center}
\vspace*{-5mm}
  \caption{The combination of multiple global descriptors (CGD) framework.
  The framework is described with ResNet-50 backbone where Stage 3 downsampling is removed.
  From the last feature map, each of $n$ global descriptor branch outputs a $k$-dimensional embedding vector, which is concatenated into the combined descriptor for ranking loss.
  Exclusively the first global descriptor is used for auxiliary classification loss where $M$ denotes the number of classes.}
\label{fig:architecture}
\vspace{-0.3em}
\end{figure*}

% 이러한 문제들을 해결하기 위해 우리는 이러한 framework를 제안하고, contribution
In this paper, we focus on how to exploit multiple global descriptors to get an ensemble effect without explicitly training multiple learners and controlling diversity among learners.
Our contribution is threefold.
(1) We propose a novel framework, the combination of multiple global descriptors (CGD), that combines multiple global descriptors which can be trained in an end-to-end manner.
It achieves an ensemble effect without any explicit ensemble model or diversity control over each global descriptor.
Moreover, the proposed framework is flexible and expandable by the global descriptor, CNN backbone, loss, and dataset.
To the best of our knowledge, we are the first to leverage multiple types of global descriptors to get a final descriptor in the image retrieval task.
(2) We investigate the effect of combining multiple global descriptors with quantitative and qualitative analysis.
Our extensive experiments demonstrate that using combined descriptor outperforms a single global descriptor because it can use different types of feature properties.
(3) The proposed framework achieves the state-of-the-art performance on CARS196~\cite{krause20133d}, CUB200-2011~\cite{wah2011caltech} (CUB200), Stanford Online Products~\cite{oh2016deep} (SOP) and In-shop Clothes~\cite{liu2016deepfashion} (In-shop) by a large margin on image retrieval tasks.

%------------------------------------------------------------------------

\section{Related Works}

%%% Global descriptor
In the recent works for image retrieval task, global descriptors based on deep CNNs have been used as off-the-shelf feature~\cite{sharif2014cnn, Babenko_2015_ICCV} over the conventional hand-crafted features such as SIFT~\cite{lowe2004distinctive}.
SPoC~\cite{Babenko_2015_ICCV} is sum pooling from the feature map which performs well mainly due to the subsequent descriptor whitening.
MAC~\cite{tolias2015particular} by max pooling is another powerful descriptor, while regional MAC~\cite{tolias2015particular} performs max pooling over regions, then sum over the regional MAC descriptors at the end.
GeM~\cite{radenovic2018fine} generalizes max and average pooling with a pooling parameter.
Other global descriptor method includes weighted sum pooling~\cite{kalantidis2016cross}, weighted GeM~\cite{wu2018weighted},  multiscale RMAC~\cite{li2017ms}, etc.

%%% Additional strategy, Attention mechanism
% Ours: no additional parameter, no bigger model
Some works~\cite{dai2018batch, gu2018attention, li2018harmonious} attempt using the additional strategy or the attention mechanism to maximize the activations of essential features on the feature map.
Dai \textit{et al.}~\cite{dai2018batch} present a strategy called batch feature erasing (BFE) to force the network to optimize the feature representation of different regions.
Li \textit{et al.}~\cite{li2018harmonious} propose a model that has soft pixel attention and hard regional attention along with simultaneous optimization of feature representations.
The downsides of adopting the additional strategy or the attention mechanism are that it can not only lead to an increase network size and training time, but also require additional parameters for training.
However, our proposed framework does not need any additional strategy or attention mechanism when it requires only a few additional parameters for training.

%%% Ensemble
% Ours: no diversity control, end-to-end training
The ensemble is a well-known technique that aims to boost performance by training multiple learners and obtains a combined result from the trained learners.
In the last decades, it is widely used in image retrieval tasks~\cite{Kim_2018_ECCV, opitz2017bier, Xuan_2018_ECCV, lin2018regional}.
Xuan \textit{et al.}~\cite{Xuan_2018_ECCV} propose a method where each embedding function is learned by randomly bagging and training labels into small subsets.
Kim \textit{et al.}~\cite{Kim_2018_ECCV} suggest an attention-based ensemble, where single feature embedding function is trained while each learner learns different attention modules.
The downside of ensemble techniques is that it leads to an increase in computational cost as the model complexity increases~\cite{zhu2018binary}, and requires additional control to yield diversity between learners~\cite{Kim_2018_ECCV, opitz2016efficient}.
However, our proposed framework takes advantage of the idea of the ensemble technique when it can be trained in an end-to-end manner with no diversity control.

\section{Proposed Framework}

We propose a simple, yet effective framework which we refer to as a CGD framework for image retrieval tasks.
It learns a combined descriptor which is generated by concatenating multiple global descriptors in an end-to-end manner.
Our proposed framework is depicted in Figure~\ref{fig:architecture}.

The proposed framework consists of a CNN backbone network and two modules.
The first module is the main module that learns an image representation, which is a combination of multiple global descriptors for a ranking loss.
Next, is an auxiliary module to fine-tune a CNN with a classification loss.
The proposed framework is trained with a final loss, which is the sum of the ranking loss from the main module and the classification loss from the auxiliary module in an end-to-end manner.

\subsection{Backbone Network}

Our proposed framework can use any CNN backbones such as BN-Inception~\cite{ioffe2015batch}, ShuffleNet-v2~\cite{ma2018shufflenet}, ResNet~\cite{he2016deep} and its variants, etc, while we use ResNet-50~\cite{he2016deep} as a baseline backbone described in Figure~\ref{fig:architecture}.
To preserve more information in the last feature map, we modify the network by discarding the down-sampling operation between Stage 3 and Stage 4~\cite{dai2018batch, wang2018learning}.
This modification gives a $14 \times 14$ sized feature map at the end for input size of $224 \times 224$, which improves the performance by containing richer information.

\subsection{Main Module: Multiple Global Descriptors} \label{sec:main_module}

The main module has multiple branches that output each image representation by using different global descriptors on the last convolutional layer.
In this paper, we use three types of the most representative global descriptors on each branch, including SPoC, MAC, and GeM.

Given an image $I$, the output of the last convolutional layer is a 3D tensor $\mathcal{X}$ of $C \times H \times W$ dimension, where $C$ is the number of feature maps.
Let $\mathcal{X}_{c}$ be the set of $H \times W$ activations for feature maps $c \in \left\{ 1 \ldots C \right\}$.
The network output consists of $C$ channels of such 2D feature maps.
Global descriptor takes $\mathcal{X}$ as input and produces a vector $f$ as output by pooling process.
Such pooling methods can be generalized as follows:
\begin{equation} \label{eq:gem}
\resizebox{.88\hsize}{!}{
    $f = [ f_1 \ldots f_c \ldots f_C ]^\top, \quad f_c = ({1 \over \left|\mathcal{X}_{c} \right|} \sum\limits_{x \in {\mathcal{X}_{c}}} x^{p_{c}})^{{1 \over p_{c}}}$
}.
\end{equation}
We define SPoC as $f^{(s)}$ when $p_{c} = 1$, MAC as $f^{(m)}$ when $p_{c} \rightarrow \infty$, and GeM as $f^{(m)}$ for the rest of the cases.
For the case of GeM, the parameter $p_{c}$ can be manually set or trained because it is differentiable, while we use fixed $p_{c}$ parameter 3 throughout the experiments.

Output feature vector $\Phi^{(a_i)}$ from the $i$-th branch is generated by dimensionality reduction through the FC layer and normalization through the $l_2$-normalization layer:
\begin{equation}
\resizebox{.7\hsize}{!}{
    $\Phi^{(a_i)} = { { W^{(i)} \cdot {f^{(a_i)}} } \over { \| W^{(i)} \cdot {f^{(a_i)}} \|_2 }}, \quad a_i \in \{ s, m, g \}, $
    % $\Phi^{(a)} = { { W \cdot {f^{(a)}} } \over { \| W \cdot {f^{(a)}} \|_2 }}, \quad a \in \{ s, m, g \}, $
}
\end{equation}
for $i \in \{ 1 \ldots n \}$, where $n$ is the number of branches, $W^i$ is the weight of the FC layer and the global descriptor $f^{(a_i)}$ can be SPoC when $a_i=s$, MAC when $a_i=m$, or GeM for $a_i=g$.

The final feature vector referred to as combined descriptor $\psi_{CGD}$ of our framework combines output feature vectors of multiple branches and performs $l_2$-normalization sequentially:
\begin{equation}
\resizebox{.8\hsize}{!}{
    $\psi_{CGD} = { \Phi^{(a_1)} \oplus \ldots \oplus \Phi^{(a_i)} \oplus \ldots \oplus \Phi^{(a_n)} \over \| { \Phi^{(a_1)} \oplus \ldots \oplus \Phi^{(a_i)} \oplus \ldots \oplus \Phi^{(a_n)}} \|_2 },$
}
\end{equation}
for $a_i \in \{ s, m, g \}$, where $\oplus$ denotes concatenation.
This combined descriptor can be trained with any ranking loss, while we use batch-hard triplet loss~\cite{HermansBeyer2017Arxiv} as a representative.

In the proposed framework, there are two advantages to combining multiple global descriptors.
First, it gives an ensemble effect with only a few additional parameters.
To get the ensemble effect while making it trainable in an end-to-end manner, our framework extracts and combines multiple global descriptors within a single CNN backbone.
Second, it automatically provides different properties for each branch's output without any diversity control.
While \cite{Kim_2018_ECCV, opitz2016efficient} propose specially designed losses to encourage diversity among learners, our framework does not require any specially designed loss to control diversity among branches.

\subsection{Auxiliary Module: Classification Loss} \label{sec:auxiliary_module}

The auxiliary module fine-tunes the CNN backbone based on the first global descriptor of the main module by using a classification loss.
It is motivated by the approach~\cite{gordo2017end}, which consists of two steps: training a CNN backbone with a classification loss and then fine-tuning the network with a triplet loss.
However, we refine their approach to have a single step for end-to-end training, while \cite{gordo2017end} has to be trained with two steps.
Training with auxiliary classification loss helps to maximize inter-class distance which makes the model to train faster and stable.

Temperature scaling~\cite{guo2017calibration, zhang2018Heated} in softmax cross-entropy loss (softmax loss), and label smoothing~\cite{szegedy2016rethinking} are proven to be helpful for the training process.
The softmax loss is defined as
\begin{equation} \label{eq:softmax}
\resizebox{.8\hsize}{!}{
    $L_{Softmax} = -{1 \over N} \sum\limits_{i=1}^N \log{ \exp{((W_{y_i}^T {f_i} + b_{y_i})/\tau)} \over {\sum\limits_{j=1}^M{\exp{((W_{j}^T {f_i} + {b_j})/\tau)}}} },$
}
\end{equation}
where $N$, $M$, and $y_i$ are the batch size, the number of classes, and the corresponding identity label of $i$-th input, respectively.
$W$, and $b$ are trainable weight, and bias, respectively.
$f$ is a global descriptor from the first branch, where $\tau$ is a temperature parameter with default value 1.
The temperature scaling with low-temperature parameter $\tau$ in the Equation~\ref{eq:softmax}, assigns a larger gradient to more challenging examples and is helpful for intra-class compact, and inter-class spread-out embedding.
The label smoothing enhances a model, thereby improves generalization by estimating the marginalized effect of a label-dropout during training.
Therefore, to prevent over-fitting, and learn better embedding, we add label smoothing and temperature scaling in the auxiliary classification loss.

\subsection{Configurations of Framework} \label{sec:configurations_of_framework}
\paragraph{Configurations}
Our proposed framework is expandable by the number of global descriptor branches, and it allows different types of networks according to the configuration of global descriptors.
As we use SPoC (S), MAC (M), GeM (G), and exclusively the first global descriptor is used for the auxiliary classification loss, we can make twelve possible configurations.
First letter in a notation is the first global descriptor to be used for the auxiliary classification loss.
For example with a configuration SMG, the first letter which is the first global descriptor S will be used for the auxiliary classification loss and all S, M, and G are concatenated to be combined descriptor for ranking loss.
Therefore, the twelve configurations are obtained as follows: S, M, G, SM, MS, SG, GS, MG, GM, SMG, MSG, GSM.

\vspace{-4mm}
\paragraph{How to Choose the Best}
As each global descriptor has different properties, the performance of each descriptor can vary by datasets~\cite{boureau2010theoretical}.
In order to find the best configuration, we evaluate every single descriptor and choose the highest and the second-highest single descriptors to use them for combination.
The number of global descriptors to combine has to be determined by the size of output dimensionality.
For a small output dimensionality, a small number of descriptors is recommended.
This rule to choose the best configuration is shown with an experiment below in Section~\ref{sec:quantitative_analysis}.

\subsection{Efficiency of Time and Memory}
Compared to previous methods of feature ensemble~\cite{ozaki2019large, chen20192nd, lin2018regional}, our proposed framework has better efficiency in terms of time and memory.
% In order to find the best combination of multiple global descriptors, both methods need to search within the number of combinations.
Because each learner of an ensemble method needs individual training and inference, ensembling $N$ number of learners with different global descriptors requires $N$ number of GPUs, and it requires post-processing step such as concatenation or normalization.
Our proposed method needs only one GPU independently of the number of global descriptors without any post-processing step because of a shared backbone.
Given limited memory resources, training and inference of a model in an end-to-end manner is beneficial in terms of time and memory.

%------------------------------------------------------------------------
\section{Experiments}
% In the following sections, we empirically show the design process of our proposed framework and investigate the effectiveness of each design part.
% And then, we show the robustness of our proposed framework on various ranking losses, CNN backbone models, and datasets by comparing our performance with current state-of-the-art metric learning approaches on commonly used image retrieval datasets.

%-------------------------------------------------------------------------
\subsection{Datasets}

We evaluate our proposed framework on image retrieval datasets including CUB200-2011~\cite{wah2011caltech}, CARS196~\cite{krause20133d}, Stanford Online Products~\cite{oh2016deep}, and In-shop Clothes~\cite{liu2016deepfashion}.
For CUB200 and CARS196, cropped images with bounding box information are used.
We follow the same training and test split as~\cite{dai2018batch, Kim_2018_ECCV, zhai2018making} for fair comparisons.

\subsection{Implementation}

All experiments are implemented using MXNet~\cite{chen2015mxnet} on a Tesla P40 GPU with 24 GB memory.
We use BN-Inception~\cite{ioffe2015batch}, ShuffleNet-v2~\cite{ma2018shufflenet}, ResNet-50~\cite{he2016deep}, SE-ResNet-50~\cite{hu2018squeeze} with ImageNet ILSVRC-2012~\cite{deng2009imagenet} pre-trained weights from MXNet GluonCV~\cite{gluoncv}.
For every experiment, we use the input size of $224 \times 224$ and the 1536-dimensional embedding, unless otherwise noted in the experiment.
In the training phase, the input image is resized to $252 \times 252$, cropped randomly to $224 \times 224$, and then flipped randomly to the horizontal.
We use an Adam~\cite{kingma2014adam} optimizer with a learning rate of 1e-4, and a step decay is used for scheduling the learning rate.
A margin of $m$ for triplet loss is 0.1, and a temperature of $\tau$ for softmax loss is 0.5, with a batch size of 128 for every experiment.
% For the batch size, 128 is used on every dataset, and for the number of instances per class, 64 is used on CARS196, and CUB200-2011, while 4 is used on In-shop Clothes, and Stanford Online Products
In the inference phase, we only resize the image by the default input size of $224 \times 224$.

%-------------------------------------------------------------------------
\subsection{Experiments for Architecture Design}
% Architecture design을 위한 실험으로, 필요한 실험 순으로 작성
% In the following sections, we empirically show the design process of the proposed framework and investigate the effectiveness of each part.

\begin{table}[t!]
\begin{center}
% \subfloat[Recall@K $\pm$ std. dev. on CARS196 dataset.]{
\begin{adjustbox}{width=.95\columnwidth,center}
\begin{tabular}{c|cccc}
\hline
\multirow{2}{*}{Loss} & \multicolumn{4}{c}{Recall@K (\%)}                \\ \cline{2-5} 
                       & 1          & 2          & 4          & 8          \\ \hline\hline
Rank                   & 86.7 $\pm$ 0.3 & 92.1 $\pm$ 0.3 & 95.3 $\pm$ 0.2 & 97.3 $\pm$ 0.1 \\
Both                   & \textbf{93.1 $\pm$ 0.1} & \textbf{96.0 $\pm$ 0.2} & \textbf{97.4 $\pm$ 0.2} & \textbf{98.3 $\pm$ 0.2} \\ \hline
\end{tabular}
\end{adjustbox}
% }

\end{center}
\caption{Recall@K $\pm$ std. dev. comparison between using only the ranking loss (Rank) and using both the classification and ranking losses (Both) on CARS196.
We report results over five runs.}
\label{table:loss}
\end{table}

\begin{table}[t!]
\begin{center}

% \subfloat[Recall@K $\pm$ std. dev. on CARS196 dataset.]{
\begin{adjustbox}{width=.95\columnwidth,center}
\begin{tabular}{c|cccc}
\hline
\multirow{2}{*}{Trick} & \multicolumn{4}{c}{Recall@K (\%)}                     \\ \cline{2-5} 
                       & 1          & 2          & 4          & 8          \\ \hline\hline
None                   & 93.1 $\pm$ 0.1 & 96.0 $\pm$ 0.2 & 97.4 $\pm$ 0.2 & 98.3 $\pm$ 0.2 \\
LS                     & 93.5 $\pm$ 0.2 & 96.1 $\pm$ 0.1 & 97.5 $\pm$ 0.1 & 98.4 $\pm$ 0.1 \\
TS                     & 94.0 $\pm$ 0.1 & 96.4 $\pm$ 0.2 & 97.8 $\pm$ 0.1 & 98.7 $\pm$ 0.1 \\
Both                   & \textbf{94.4 $\pm$ 0.2} & \textbf{96.8 $\pm$ 0.0} & \textbf{98.0 $\pm$ 0.0} & \textbf{98.8 $\pm$ 0.1} \\ \hline
\end{tabular}
\end{adjustbox}
% }

\end{center}
\caption{Recall@K $\pm$ std. dev. among the baseline `no tricks' (None), label smoothing (LS), temperature scaling (TS), and `both tricks' (Both) on CARS196.
We report results over five runs.}
\label{table:tricks}
\end{table}

\subsubsection{Training Classification and Ranking Loss Jointly}
% Classification loss를 사용했을때 얻는 효과

\paragraph{Auxiliary Classification Loss} \label{sec:ranking_loss_and_classification_loss}
Our proposed framework is trained by a ranking loss with an auxiliary classification loss from a descriptor of the first branch.
We compare the performance between using the ranking loss exclusively, and the ranking loss with the auxiliary classification loss on CARS196 in the Table~\ref{table:loss}.
In this experiment, we do not apply label smoothing, and temperature scaling on the auxiliary classification loss in every case.
It shows that using both losses provides higher performance than using ranking loss exclusively.
Classification loss focuses on clustering each class into a close embedding space on a categorical level.
Ranking loss focuses on gathering samples in the same class and making a distance between samples from the different classes in the instance level.
Therefore, training the ranking loss with the auxiliary classification loss jointly gives better optimization for categorical, and fine-grained feature embedding.

\vspace{-4mm}
\paragraph{Label Smoothing and Temperature Scaling} \label{sec:label_smoothing_and_temperature_scaling}
As mentioned in Section~\ref{sec:auxiliary_module}, label smoothing, and temperature scaling are proven to be helpful to learn better embedding for the classification loss.
We investigate if it can be applied when a model is trained with both the ranking loss and the auxiliary classification loss.
We show the performance appraisal of the `no tricks', the label smoothing, the temperature scaling with temperature term 0.5, and `both tricks' on the auxiliary classification loss in Table~\ref{table:tricks}.
The experiment is performed on the ResNet-50~\cite{he2016deep} backbone with the configuration SM.
It shows that each label smoothing and temperature scaling improves the performance compared to the `no tricks'.
Moreover, applying `both tricks' together stacks up each performance boost, and gives the best performance.

\begin{figure}[t!]
\begin{center}
\subfloat[Architecture type A.]{
    \label{fig:typeA}
    \includegraphics[clip,width=.8\columnwidth]{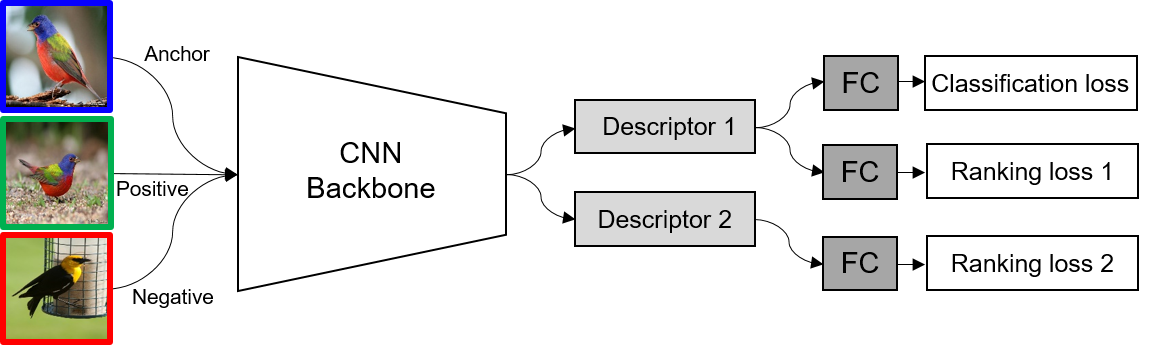}
}
\vspace{0.1em}
\subfloat[Architecture type B.]{
    \label{fig:typeB}
    \includegraphics[clip,width=.9\columnwidth]{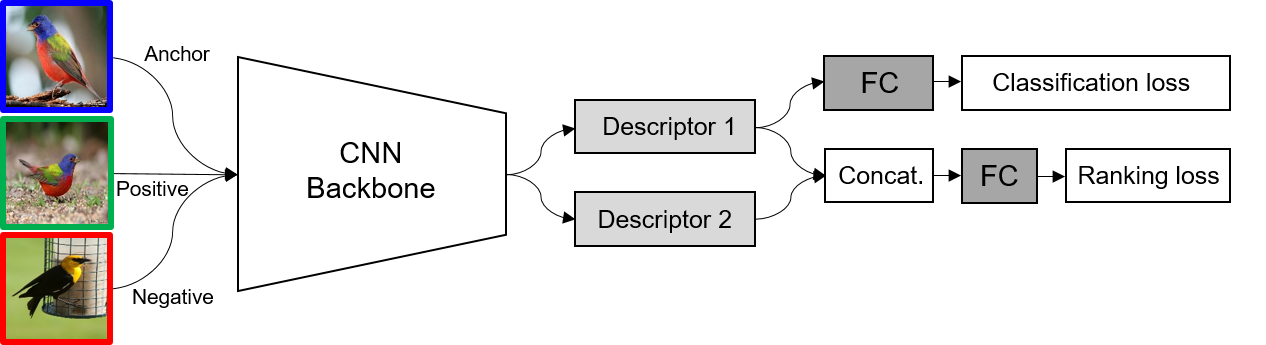}
}
\end{center}
   \caption{Different architecture types for training multiple global descriptors.}
\label{fig:architecture_type}
\vspace{-0.3em}
\end{figure}

\subsubsection{Combining Multiple Global Descriptors}

\paragraph{Position of Combination} \label{sec:position_of_combination}
As our proposed framework uses multiple global descriptors, we perform experiments with different positions of a combination of multiple global descriptors to choose the best architecture.
Architecture type A in the Figure~\ref{fig:typeA} trains each global descriptor with individual ranking loss, and then combines them at the inference phase as in~\cite{Kim_2018_ECCV}, while they use the same global descriptor for every branch and do not use classification loss.
Architecture type B in the Figure~\ref{fig:typeB} combines the raw output of global descriptors and train it with a single ranking loss, similar to studies of~\cite{shen2017learning, suvisual}, while they do not use multiple global descriptors.
Also, our proposed framework combines multiple global descriptors after the FC layers and $l2$-normalization as described in Figure~\ref{fig:architecture}.
As shown in Table~\ref{table:architecture_type}, the proposed position of the combination presents the best performance over the architecture type A and type B.
The reason is that CGD can maintain properties and diversities of each feature vector from multiple branches.
In contrast, the final embedding of type A in the training phase is different from that of the inference phase, and the final embedding of type B loses each property of the global descriptors because they are mixed up by FC layer after concatenation.

\begin{table}[t!]
\begin{center}
\begin{adjustbox}{width=0.95\columnwidth,center}
\begin{tabular}{c|cccc}
\hline
\multirow{2}{*}{Type}   & \multicolumn{4}{c}{Recall@K (\%)} \\ \cline{2-5} 
                        & 1                     & 2                     & 4                     & 8                     \\
                       \hline\hline
A                       & 74.6 $\pm$ 0.4	        & 83.5 $\pm$ 0.4	        & 89.8 $\pm$ 0.3	        & \textbf{94.0 $\pm$ 0.2}   \\
B                       & 73.7 $\pm$ 0.3	        & 82.6 $\pm$ 0.3	        & 89.2 $\pm$ 0.2	        & 93.5 $\pm$ 0.2            \\
CGD                 & \textbf{75.3 $\pm$ 0.5}	& \textbf{83.9 $\pm$ 0.3}	& \textbf{89.9 $\pm$ 0.3}	& \textbf{94.0 $\pm$ 0.3}   \\ \hline
\end{tabular}
\end{adjustbox}
\end{center}
\caption{Recall@K $\pm$ std. dev. among architecture type A, type B, and the proposed framework with the configuration SM on CUB200-2011.
We report results over five runs.}
\label{table:architecture_type}
\end{table}

\vspace{-4mm}
\paragraph{Method of Combination}
% Combination 방법으로 가능한 concat 와 sum 에 대한 설명, summation에 대한 reference가 있으면 좋겠다.
In terms of the combination method, concatenation, and summation of multiple descriptors are proven to enhance performance in~\cite{Kim_2018_ECCV, shen2017learning, suvisual, tolias2015particular, dai2018batch}.
Therefore, we compare two combination methods to choose the best.
As shown in Table~\ref{table:combination}, concatenation of multiple global descriptors gives better performance compared to their summation.
This also indicates the importance of preserving each property, and diversity from multiple global descriptors, as the summation mix activations of each global descriptor up, while the concatenation maintains them.

\begin{table}[t!]
\begin{center}
\begin{adjustbox}{width=0.96\columnwidth,center}
\begin{tabular}{c|cccc}
\hline
\multirow{2}{*}{Comb.} & \multicolumn{4}{c}{Recall@K (\%)} \\ \cline{2-5} 
                        & 1                     & 2                     & 4                     & 8                     \\
                       \hline\hline
Sum                     & 73.8 $\pm$ 0.5	        & 82.9 $\pm$ 0.4	        & 89.4 $\pm$ 0.3	        & 93.7 $\pm$ 0.1            \\ 
Concat	                & \textbf{75.3 $\pm$ 0.5}	& \textbf{83.9 $\pm$ 0.3}	& \textbf{89.9 $\pm$ 0.3}	& \textbf{94.0 $\pm$ 0.3}   \\ \hline
\end{tabular}
\end{adjustbox}
\end{center}
\caption{Recall@K $\pm$ std. dev. comparison by combination method with the configuration SM on CUB200-2011.
We report results over five runs.}
\label{table:combination}
\vspace{-0.3em}
\end{table}

\subsection{Effectiveness of Combined Descriptor} \label{sec:effectiveness_of_combined_descriptor}
\subsubsection{Quantitative Analysis} \label{sec:quantitative_analysis}

\begin{figure*}[h!t!] %
\vspace{-2.2em}
\centering
\subfloat[Recall@1 (\%) on CARS196.]{%
\includegraphics[width=0.45\textwidth]{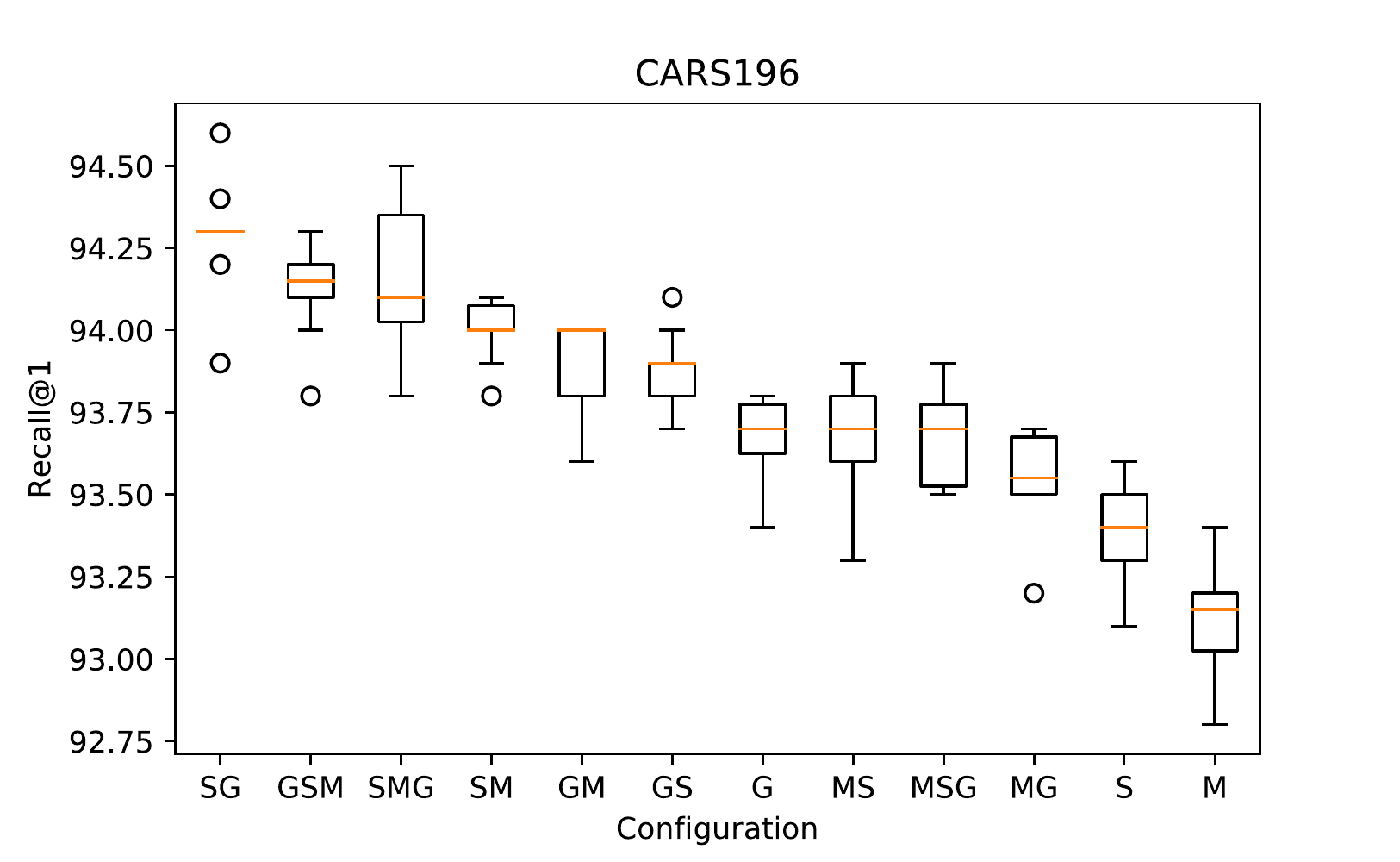}%
\label{fig:cars_config}%
}\hfil
\subfloat[Recall@1 (\%) on CUB200-2011.]{%
\includegraphics[width=0.45\textwidth]{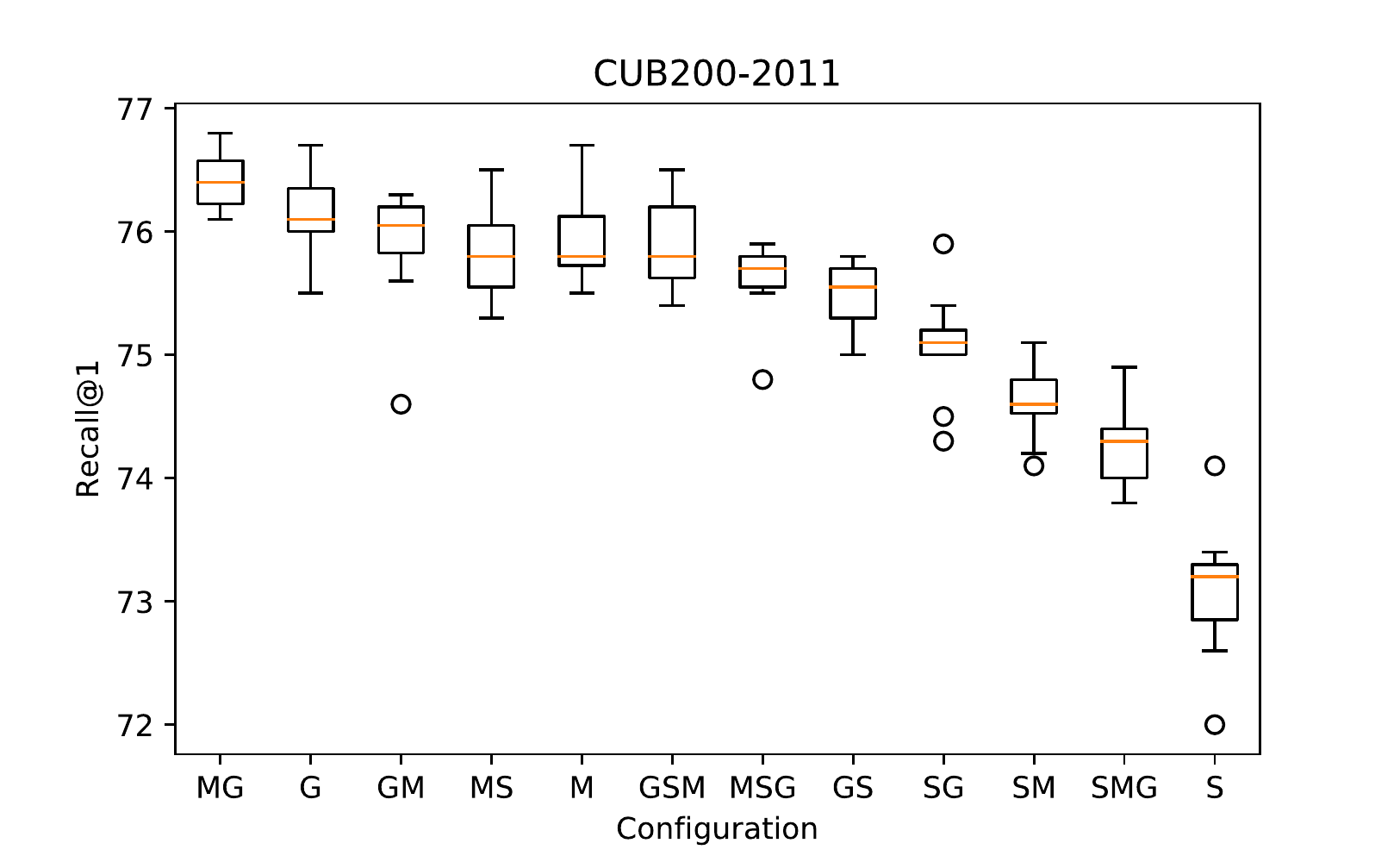}%
\label{fig:cub_config}%
}
\vspace{-1em}
\subfloat[Recall@1 (\%) on Stanford Online Products.]{%
\includegraphics[width=0.45\textwidth]{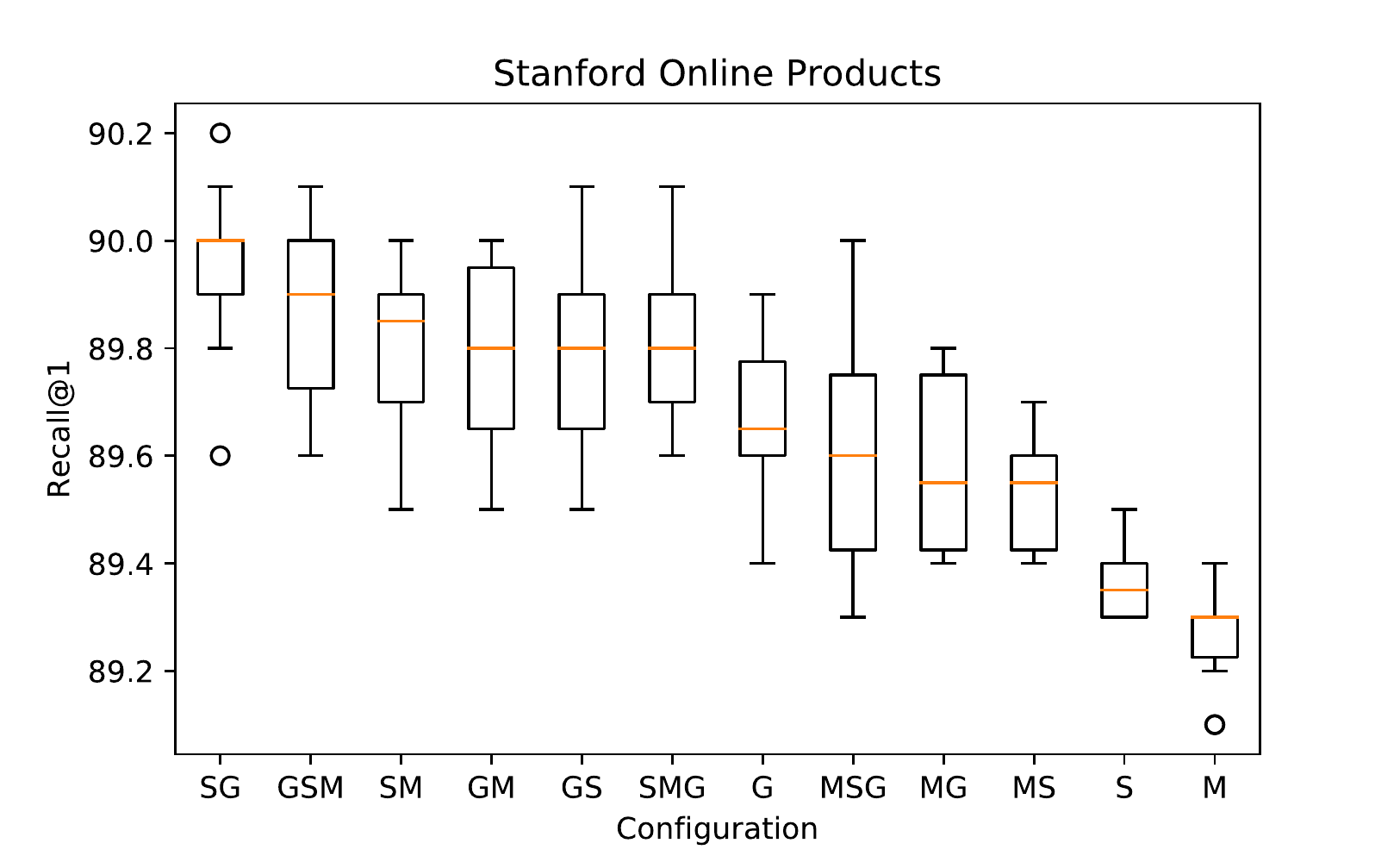}%
\label{fig:sop_config}%
}\hfil
\subfloat[Recall@1 (\%) on In-shop Clothes.]{%
\includegraphics[width=0.45\textwidth]{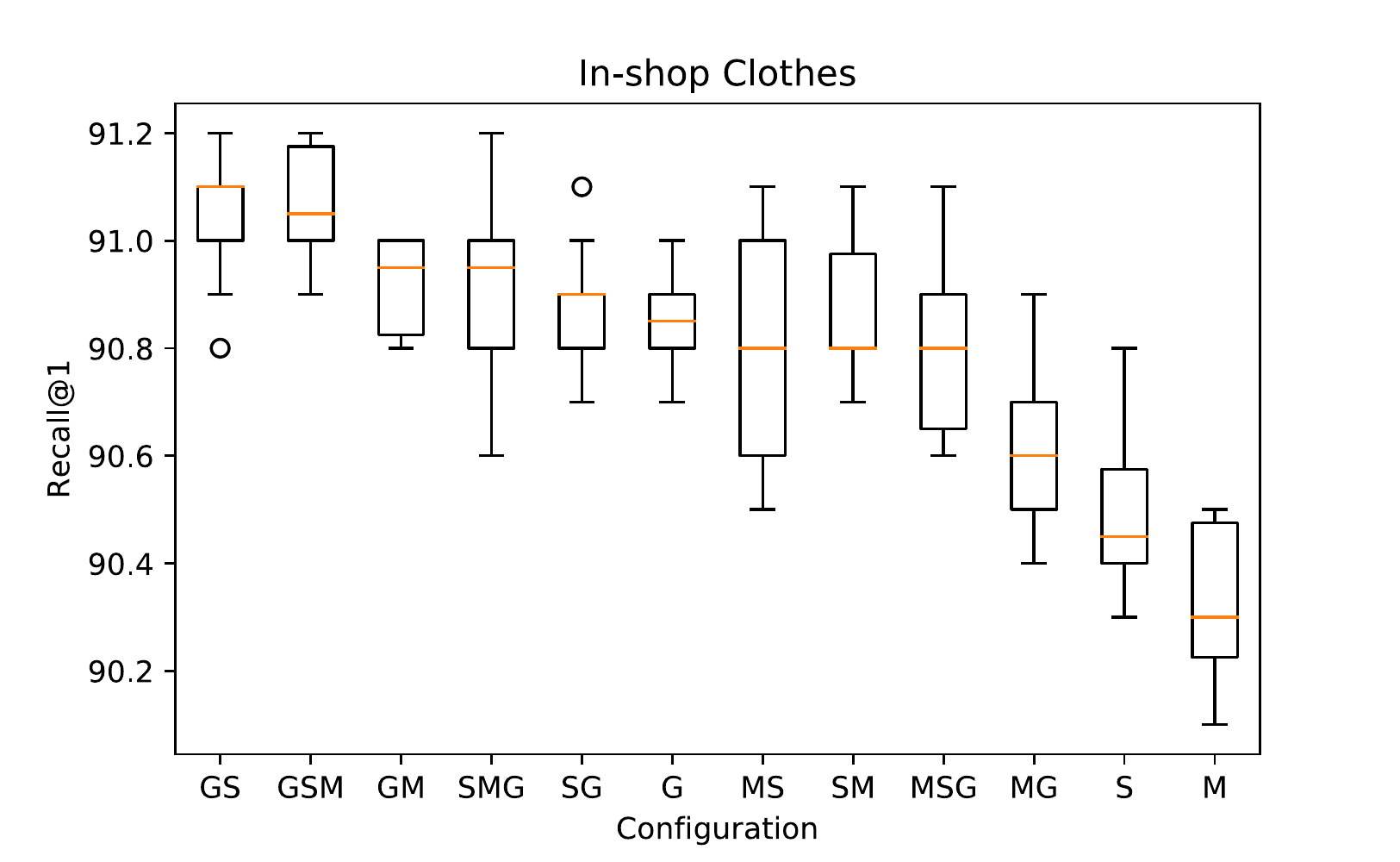}%
\label{fig:inshop_config}%
}
\vspace{1em}
\caption{Performance of different configurations of our proposed framework.
For the faster experiments on SOP, we use a mini-test set by sampling a hundred instances per class.
Due to the uncertainty of deep learning model, we report results over ten runs with box plots.
}
\label{fig:configuration}
\vspace{-0.3em}
\end{figure*}

The core of our proposed framework is exploiting multiple global descriptors.
As we defined in Section~\ref{sec:configurations_of_framework}, we conduct experiments with twelve possible configurations on each image retrieval dataset.
% , where the framework uses the temperature scaling exclusively in the auxiliary classification loss.
In the Figure~\ref{fig:configuration}, majorities of combined descriptors outperform over than single global descriptors.
% For the CUB200, the single global descriptors G and M show relatively high performance while the best performance configuration is still combined descriptor MG.
Moreover, the best configuration is a combination of the highest and the second-highest single descriptors, which we will use this pattern to find the best configuration, as mentioned in Section~\ref{sec:configurations_of_framework}.
% The performance can be varied by the properties of datasets, the feature used for the classification loss, the size of the input and the output dimension, etc.
While the performance of each descriptor is varied by the properties of datasets, the main essence is that exploiting multiple global descriptors gives performance boost compared to single global descriptors.

Table~\ref{table:individual_descriptor} shows the performance of individual global descriptors before combining operation and how much performance gain they can produce after the operation.
Every combined descriptor have 1536-dimensional embedding vector, while individual descriptor has 1536-dimensional embedding vector for S, M, G, 768-dimensional embedding vector for SM, MS, SG, GS, MG, GM, and 512-dimensional embedding vector for SMG, MSG, GSM.
Having a larger embedding dimension usually gives better performances.
However, if the performance difference is not much between a large embedding and a small embedding, it may be preferable to use multiple small embeddings from different global descriptors.
For example, as individual descriptor GeM from SG with 768 embedding dimensions has similar performance with a single descriptor G with 1536 embedding dimensions, SG gets a significant performance boost by combining different features of SPoC, and GeM.

\begin{table}[h]
\begin{center}
\begin{adjustbox}{width=0.85\columnwidth,center}
\begin{tabular}{c|c|ccc|c}
\hline
\multirow{2}{*}{Config.} & \multirow{2}{*}{\begin{tabular}[c]{@{}c@{}}Combined\\ (1536-dim.)\end{tabular}} & \multicolumn{4}{c}{Individual Descriptor} \\ \cline{3-6} 
                        &                                                                                 & SPoC & MAC  & GeM  & Dim.                  \\ \hline\hline
S                       & -                                                                               & \textbf{93.8} & -    & -    & \multirow{3}{*}{1536} \\
M                       & -                                                                               & -    & 93.6 & -    &                       \\
G                       & -                                                                               & -    & -    & 93.9 &                       \\ \hline
SM                      & 94.3                                                                            & 92.5 & 93.6 & -    & \multirow{6}{*}{768}  \\
MS                      & 94.0                                                                            & 93.2 & 93.5 & -    &                       \\
SG                      & \textbf{94.5}                                                                            & 93.0 & -    & \textbf{94.0} &                       \\
GS                      & 94.2                                                                            & 93.5 & -    & 93.9 &                       \\
MG                      & 93.9                                                                            & -    & 93.4 & 93.3 &                       \\
GM                      & 94.2                                                                            & -    & \textbf{93.9} & 93.3 &                       \\ \hline
SMG                     & 94.2                                                                            & 92.2 & 93.0 & 93.0 & \multirow{3}{*}{512}  \\
MSG                     & 94.4                                                                            & 92.7 & 93.0 & 93.8 &                       \\
GSM                     & 94.0                                                                            & 92.7 & 93.2 & 93.0 &                       \\ \hline
\end{tabular}
\end{adjustbox}
\end{center}
\caption{Recall@1 (\%) of combined descriptor, and their individual descriptor on CARS196.
Each individual descriptor is an output feature vector of each branch right before the concatenating operation.
The combined descriptor is the final feature vector of the proposed framework.
We report median values from ten runs.}
\label{table:individual_descriptor}
\vspace{-0.7em}
\end{table}

\subsubsection{Qualitative Analysis} \label{sec:qualitative_analysis}

\begin{figure*}[h!t!]
\vspace{-0.5em}
\begin{center}
\includegraphics[width=0.9\textwidth]{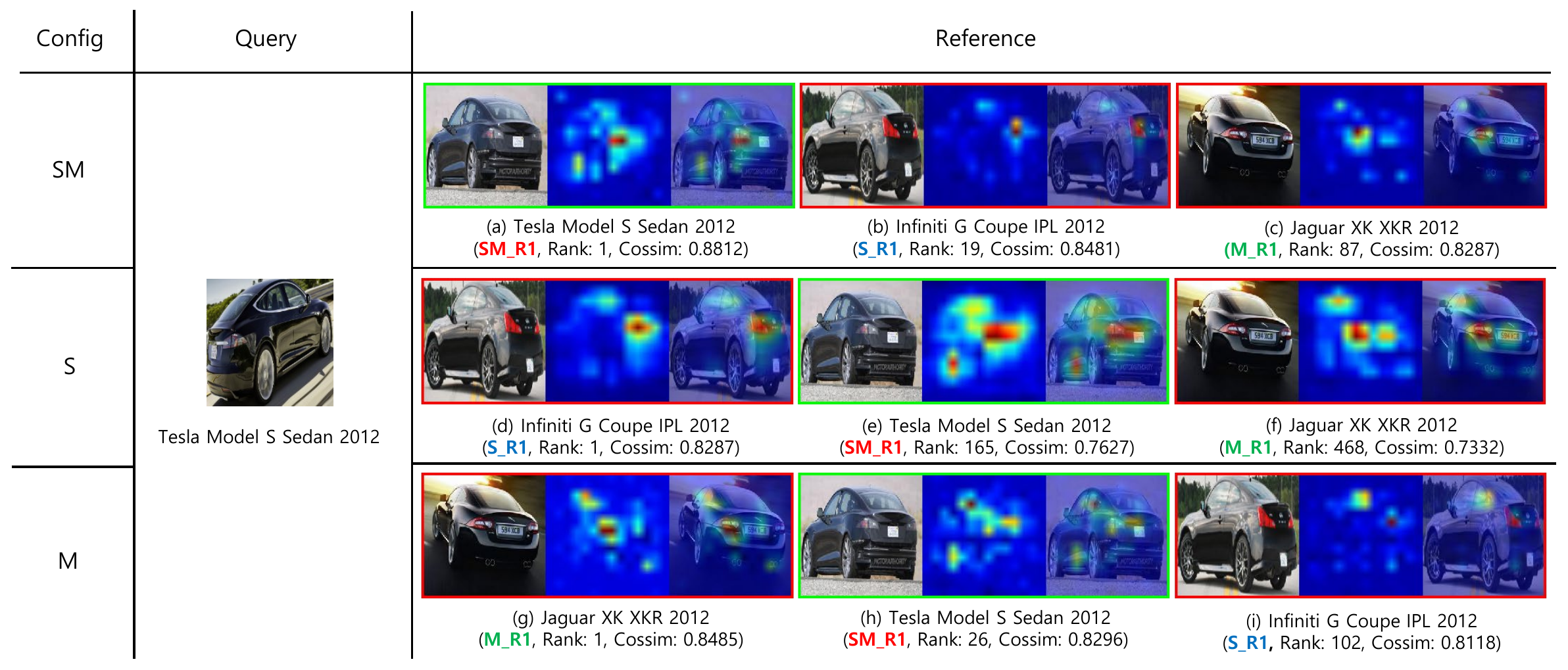}
\end{center}
   \caption{Spatial similarity visualizations on CARS196.
%   Each heatmap is drawn with respect to the given query embedding, and it shows how spatial regions contribute to the pairwise similarity.
   Heatmaps show the contribution of each region toward pairwise similarity computation.
   For each configuration SM, S, and M, we visualize the top 1 retrieved image (a), (d), and (g), respectively, from the same query image.
   These images are denoted by {\color{red}\textbf{SM\_R1}}, {\color{blue}\textbf{S\_R1}}, and {\color{OliveGreen}\textbf{M\_R1}}.
   They are used to visualize on different configurations (b, c, e, f, h, i) so that we can see the rank changes among the configurations.
   The green box indicates that the image is in the same class, while the red box indicates that the image is in a different class.
   ``Cossim'' denotes cosine similarity.
   }
\label{fig:vis}
\vspace{-0.3em}
\end{figure*}

% Visualization으로 proposed method의 효과 주장
% Abby \textit{et al.}~\cite{stylianouSimVis2019} propose a visualization tools to process the embedding generated by deep similarity networks.
A visualization tool proposed in~\cite{stylianouSimVis2019} highlights the regions of images that contribute the most to pairwise similarity.
We modify this work for our framework to see how much each region of an image contributes to the similarity for each final embedding.
Figure~\ref{fig:vis} shows a visualization of topmost (Recall@1) retrieved image of each configuration on the same query.

As mentioned in~\cite{stylianouSimVis2019}, the regions of similarity are large in the configuration S, while the configuration M has more focused regions of similarity.
The configuration SM seems to have the similarity regions mixed with the configuration S, and M.
SPoC tends to see the overall information when it lacks discriminability because they average the high activated outputs by non-active outputs~\cite{hoang2017selective}.
MAC is preferable to retain the high activation when it is only powerful for sparse features~\cite{boureau2010theoretical}.
However, the configuration SM, which has both properties of SPoC, and MAC seems to keep the overall information, and also retain the discriminative regions.
This property of the configuration SM pushes up the Figure {\color{red}4e} at the rank of 165 in the configuration S and Figure {\color{red}4h} at the rank of 26 in the configuration M into the rank of 1 as Figure {\color{red}4a}.
Following this experiment, combining multiple global descriptors allows the use of different properties of the global descriptors that can help to compute similarity.

\subsection{Flexibility of CGD Framework}
% Robustness를 주장
% In this section, we empirically show the robustness of our proposed network by applying different types of ranking losses, convolutional neural network backbones, and datasets.
% Moreover, we compare the results of our approach with current state-of-the-art methods on each image retrieval datasets.

% \subsubsection{Ranking Loss} \label{sec:ranking loss}
\paragraph{Ranking Loss}
% 여러 Ranking loss를 사용했을때 robust하다
Table~\ref{table:losses} shows that CGD framework can use various ranking losses, such as soft-margin or batch-hard triplet loss~\cite{HermansBeyer2017Arxiv}, HAP2S loss~\cite{yu2018hard}, and weighted sampling margin loss~\cite{wu2017sampling}.
We compare the performance of the configuration S as a baseline for single global descriptor, and SM for multiple global descriptors using these losses.
Coefficient $\alpha$ for HAP2S P loss is set to 10, coefficient $\sigma$ for HAP2S E loss is set to 0.5, and margin $\alpha$ and boundary $\beta$ for margin loss are fixed at 0.1 and 1.2, respectively.
In every case, the performance of the configuration SM is better than S, which shows that our framework is flexible in applying various losses.

\begin{table}[t!]
\begin{center}

% \subfloat[Recall@K $\pm$ std. dev. on CARS196 dataset.]{
\begin{adjustbox}{width=.5\textwidth,center}
\begin{tabular}{c|cccc}
\hline
\multirow{2}{*}{Config. + Loss}                             & \multicolumn{4}{c}{Recall@K (\%)}                 \\ \cline{2-5}
                                                           & 1          & 2          & 4          & 8          \\ \hline\hline
S + Triplet$^\dagger$          & 93.8 $\pm$ 0.2 & 96.5 $\pm$ 0.1 & 97.8 $\pm$ 0.1 & 98.7 $\pm$ 0.1 \\
SM + Triplet$^\dagger$         & \textbf{94.3 $\pm$ 0.2} & \textbf{96.8 $\pm$ 0.2} & \textbf{98.1 $\pm$ 0.1} & \textbf{98.9 $\pm$ 0.1} \\ \hline
S + Triplet$^\ddagger$         & 89.5 $\pm$ 0.1 & 94.2 $\pm$ 0.1 & 96.7 $\pm$ 0.1 & 98.2 $\pm$ 0.1 \\
SM + Triplet$^\ddagger$        & \textbf{90.4 $\pm$ 0.2} & \textbf{94.8 $\pm$ 0.2} & \textbf{97.1 $\pm$ 0.1} & \textbf{98.5 $\pm$ 0.0} \\ \hline
S + HAP2S E                              & 93.8 $\pm$ 0.3 & 96.6 $\pm$ 0.1 & 98.0 $\pm$ 0.1 & 98.8 $\pm$ 0.1 \\
SM + HAP2S E                             & \textbf{94.6 $\pm$ 0.1} & \textbf{97.0 $\pm$ 0.1} & \textbf{98.2 $\pm$ 0.1} & \textbf{98.9 $\pm$ 0.0} \\ \hline
S + HAP2S P                              & 94.4 $\pm$ 0.1 & 96.9 $\pm$ 0.1 & \textbf{98.2 $\pm$ 0.1} & \textbf{98.9 $\pm$ 0.0} \\
SM + HAP2S P                             & \textbf{95.0 $\pm$ 0.1} & \textbf{97.2 $\pm$ 0.1} & \textbf{98.2 $\pm$ 0.1} & \textbf{98.9 $\pm$ 0.1} \\ \hline
S + Margin                           & 92.8 $\pm$ 0.2 & 95.7 $\pm$ 0.1 & 97.3 $\pm$ 0.1 & 98.3 $\pm$ 0.1 \\
SM + Margin                          & \textbf{93.9 $\pm$ 0.2} & \textbf{96.4 $\pm$ 0.1} & \textbf{97.7 $\pm$ 0.1} & \textbf{98.6 $\pm$ 0.1} \\ \hline
% \footnotesize{$^a$ The smallest spatial unit is county, $^b$ more details in appendix A}
\end{tabular}
\end{adjustbox}
% }

\end{center}

\caption{Recall@K $\pm$ std. dev. of the single global descriptor S as a baseline and the combined descriptor SM with various ranking losses on CARS196. $^\dagger$ denotes the batch-hard triplet, and $^\ddagger$ denotes the soft-margin hard triplet.
We report results over five runs.}
\label{table:losses}
\vspace{-0.3em}
\end{table}

\vspace{-4mm}
\paragraph{Backbone}
% \subsubsection{Backbone} \label{sec:backbone}
% 여러 backbone을 사용했을때 robust하다. (SOTA table 에서 함께 주장)
Our framework can use different types of CNN backbone.
We perform experiments on image retrieval datasets with various CNN backbone: BN-Inception~\cite{ioffe2015batch}, ShuffleNet-v2~\cite{ma2018shufflenet}, ResNet-50~\cite{he2016deep}, and SE-ResNet-50~\cite{hu2018squeeze}.
In Table~\ref{table:sota1} and Table~\ref{table:sota2}, each value of the same color indicates the same CNN backbone and embedding dimension, which demonstrates that the CGD framework outperforms existing models with the same backbone.
Additional experiments with ShuffleNet-v2 presents a reasonable performance even though it is a compact network.
Other experiments with SE-ResNet-50 provide the best performance among all as the backbone is very powerful.
% 더 성능이 낮은 backbone이 성능이 좋은 backbone을 이기면 언급해주자.

\vspace{-4mm}
\paragraph{Dataset: Comparison with State-of-the-Art}
% \subsubsection{Dataset: Comparison with State-of-the-Art}
% 여러 dataset을 사용했을때 robust하다.

\begin{table*}[h!t]
\begin{center}
\subfloat[Recall@K (\%) on CUB200-2011 (cropped) and CARS196 (cropped). CGD (MG/SG) denotes that the configuration MG is used for CUB200-2011 and SG is used for CARS196 on the proposed CGD framework.]{
    \begin{adjustbox}{width=0.77\textwidth,center}
    \begin{tabular}{c|c|c|cccc|cccc}
    \hline
    \multirow{2}{*}{Model} & \multirow{2}{*}{Backbone} & \multirow{2}{*}{Dim} & \multicolumn{4}{c|}{CUB200} & \multicolumn{4}{c}{CARS196} \\ \cline{4-11} 
                           &                           &                      & 1     & 2     & 4    & 8    & 1     & 2     & 4    & 8    \\ \hline\hline
    Facility~\cite{oh2017deep} & BN-Inception               & 64                   & 48.2  & 61.4  & 71.8 & 81.9 & 58.1  & 70.6  & 80.3 & 87.8 \\
    Proxy-NCA~\cite{movshovitz2017no} & \color{Plum}BN-Inception               & \color{Plum}64                   & {\color{Plum}49.2}  & {\color{Plum}61.9}  & {\color{Plum}67.9} & {\color{Plum}72.4} & {\color{Plum}73.2}  & {\color{Plum}82.4}  & {\color{Plum}86.4} & {\color{Plum}88.7} \\
    % A-BIER~\cite{opitz2018deep} & GoogLeNet                 & 512                  & 57.5  & 68.7  & 78.3 & 86.2 & 82.0  & 89.0  & 93.2 & 96.1 \\
    HTL~\cite{ge2018deep}  & \color{OliveGreen}BN-Inception               & \color{OliveGreen}512                  & {\color{OliveGreen}57.1}  & {\color{OliveGreen}68.8}  & {\color{OliveGreen}78.7} & {\color{OliveGreen}86.5} & {\color{OliveGreen}81.4}  & {\color{OliveGreen}88.0}  & {\color{OliveGreen}92.7} & {\color{OliveGreen}95.7} \\
    Margin~\cite{wu2017sampling} & \color{blue}ResNet-50                  & \color{blue}128                  & {\color{blue}63.9}  & {\color{blue}75.3}  & {\color{blue}84.4} & {\color{blue}90.6} & {\color{blue}86.9}  & {\color{blue}92.7}  & {\color{blue}95.6} & {\color{blue}97.6} \\
    ABE-8~\cite{Kim_2018_ECCV} & GoogleNet$^\ddagger$        & 512                  & 70.6  & 79.8  & 86.9 & 92.2 & 93.0  & 95.9  & 97.5 & 98.5 \\
    BFE$^\dagger$~\cite{dai2018batch} & \color{red}ResNet-50$^\ddagger$         & \color{red}1536                 & {\color{red}74.1}  & {\color{red}83.6}  & {\color{red}89.8} & {\color{red}93.6} & {\color{red}94.3}  & {\color{red}96.8}  & {\color{red}\textbf{98.3}} & {\color{red}\textbf{98.9}} \\ \hline
    CGD (MG/SG)             & \color{Plum}BN-Inception               & \color{Plum}64                   & {\color{Plum}{\textbf{61.8}}}  & \color{Plum}{\textbf{73.2}}  & \color{Plum}{\textbf{82.5}} & \color{Plum}{\textbf{89.5}} & \color{Plum}{\textbf{85.7}}  & \color{Plum}{\textbf{91.7}}  & \color{Plum}{\textbf{95.1}} & \color{Plum}{\textbf{97.3}} \\
    CGD (MG/SG)             & \color{OliveGreen}BN-Inception               & \color{OliveGreen}512                  & {\color{OliveGreen}\textbf{71.9}}  & {\color{OliveGreen}\textbf{81.1}}  & {\color{OliveGreen}\textbf{88.2}} & {\color{OliveGreen}\textbf{92.9}} & {\color{OliveGreen}\textbf{91.2}}  & {\color{OliveGreen}\textbf{95.1}}  & {\color{OliveGreen}\textbf{97.0}} & {\color{OliveGreen}\textbf{98.0}} \\
    CGD (MG/SG)             & \color{blue}ResNet-50                  & \color{blue}128                  & {\color{blue}\textbf{67.6}}  & {\color{blue}\textbf{78.1}}  & {\color{blue}\textbf{86.3}} & {\color{blue}\textbf{91.9}} & {\color{blue}\textbf{90.1}}  & {\color{blue}\textbf{94.3}}  & {\color{blue}\textbf{96.6}} & {\color{blue}\textbf{98.1}} \\
    CGD (MG/SG)             & \color{red}ResNet-50$^\ddagger$         & \color{red}1536                 & {\color{red}\textbf{76.8}}  & {\color{red}\textbf{\textbf{84.8}}}  & {\color{red}\textbf{90.6}} & {\color{red}\textbf{94.3}} & {\color{red}\textbf{94.7}}  & {\color{red}\textbf{97.0}}  & {\color{red}98.1} & {\color{red}\textbf{98.9}} \\
    CGD (MG/SG)             & ShuffleNet-v2             & 1536                 & 66.4  & 76.5  & 84.8 & 91.2 & 86.1  & 91.9  & 94.9 & 97.1 \\
    CGD (MG/SG)             & SE-ResNet-50$^\ddagger$      & 1536                 & \textbf{79.2}  & \textbf{86.6}  & \textbf{92.0} & \textbf{95.1} & \textbf{94.8}  & \textbf{97.1}  & 98.2 & 98.8 \\ \hline
    \end{tabular}
    \end{adjustbox}
    \label{table:sota1}
}

\vspace{0.1em}
\subfloat[Recall@K (\%) on Stanford Online Products and In-shop Clothes. CGD (SG/GS) denotes that the configuration SG is used for Stanford Online Products and GS is used for In-shop Clothes on the proposed CGD framework.]{
    \begin{adjustbox}{width=0.85\textwidth,center}
    \begin{tabular}{c|c|c|cccc|cccccc}
    \hline
    \multirow{2}{*}{Model} & \multirow{2}{*}{Backbone} & \multirow{2}{*}{Dim} & \multicolumn{4}{c|}{SOP}   & \multicolumn{6}{c}{In-shop}            \\ \cline{4-13} 
                           &                           &                      & 1    & 10   & 100  & 1000 & 1    & 10   & 20   & 30   & 40   & 50   \\ \hline\hline
    Facility~\cite{oh2017deep} & \color{Plum}BN-Inception               & \color{Plum}64                  & {\color{Plum}67.0} & {\color{Plum}83.7} & {\color{Plum}93.2} & -    & -    & -    & -    & -    & -    &      \\
    % A-BIER~\cite{opitz2018deep} & GoogleNet                 & 512                  & 74.2 & 86.9 & 94.0 & 97.8 & 83.1 & 95.1 & 96.9 & 97.5 & 97.8 & 98.0 \\
    HTL~\cite{ge2018deep} & \color{OliveGreen}BN-Inception               & \color{OliveGreen}512                  & {\color{OliveGreen}74.8} & {\color{OliveGreen}88.3} & {\color{OliveGreen}94.8} & {\color{OliveGreen}98.4} & -    & -    & -    & -    & -    & -    \\
    HTL~\cite{ge2018deep} & \color{OliveGreen}BN-Inception               & \color{OliveGreen}128                  & -    & -    & -    & -    & {\color{OliveGreen}80.9} & {\color{OliveGreen}94.3} & {\color{OliveGreen}95.8} & {\color{OliveGreen}97.2} & {\color{OliveGreen}97.4} & {\color{OliveGreen}97.8} \\
    Margin~\cite{wu2017sampling} & \color{blue}ResNet-50                  & \color{blue}128                  & {\color{blue}72.7} & {\color{blue}86.2} & {\color{blue}93.8} & {\color{blue}98.0} & -    & -    & -    & -    & -    & -    \\
    ABE-8~\cite{Kim_2018_ECCV} & GoogleNet$^\ddagger$        & 512                  & 76.3 & 88.4 & 94.8 & 98.2 & 87.3 & 96.7 & 97.9 & 98.2 & 98.5 & 98.7 \\
    BFE$^\dagger$~\cite{dai2018batch} & \color{red}ResNet-50$^\ddagger$         & \color{red}1536                 & {\color{red}83.0} & {\color{red}93.3} & {\color{red}97.3} & {\color{red}\textbf{99.2}} & {\color{red}89.1} & {\color{red}96.3} & {\color{red}97.6} & {\color{red}98.5} & {\color{red}\textbf{99.1}} & -    \\ \hline
    CGD (SG/GS)             & \color{Plum}BN-Inception               & \color{Plum}64                   & {\color{Plum}\textbf{75.6}} & {\color{Plum}\textbf{89.0}} & {\color{Plum}\textbf{95.5}} & {\color{Plum}\textbf{98.6}} & 86.6 & 96.3 & 97.4 & 97.9 & 98.2 & 98.4 \\
    CGD (SG/\space\space-\space\space)& \color{OliveGreen}BN-Inception     & \color{OliveGreen}512                  & {\color{OliveGreen}\textbf{80.5}} & {\color{OliveGreen}\textbf{92.1}} & {\color{OliveGreen}\textbf{96.7}} & {\color{OliveGreen}\textbf{98.9}} & -    & -    & -    & -    & -    & -    \\
    CGD (\space\space-\space\space/GS)& \color{OliveGreen}BN-Inception     & \color{OliveGreen}128                  & -    & -    & -    & -    & {\color{OliveGreen}\textbf{88.5}} & {\color{OliveGreen}\textbf{97.1}} & {\color{OliveGreen}\textbf{98.0}} & {\color{OliveGreen}\textbf{98.5}} & {\color{OliveGreen}\textbf{98.8}} & {\color{OliveGreen}\textbf{98.9}}   \\
    CGD (SG/GS)             & \color{blue}ResNet-50                  & \color{blue}128                  & {\color{blue}\textbf{81.0}} & {\color{blue}\textbf{92.2}} & {\color{blue}\textbf{96.8}} & {\color{blue}\textbf{99.1}} & 88.4 & 97.2 & 98.1 & 98.4 & 98.7 & 98.8 \\
    CGD (SG/GS)             & \color{red}ResNet-50$^\ddagger$         & \color{red}1536                 & {\color{red}\textbf{83.9}} & {\color{red}\textbf{93.8}} & {\color{red}\textbf{97.5}} & {\color{red}\textbf{99.2}} & {\color{red}\textbf{90.9}} & {\color{red}\textbf{98.0}} & {\color{red}\textbf{98.7}} & {\color{red}\textbf{99.0}} & {\color{red}\textbf{99.1}} & {\color{red}\textbf{99.2}} \\
    CGD (SG/GS)             & ShuffleNet-v2             & 1536                 & 78.7   & 90.9   & 96.1   & 98.8  & 86.1 & 96.9 & 97.8 &  98.4 & 98.6 & 98.7   \\
    CGD (SG/GS)             & SE-ResNet-50$^\ddagger$      & 1536                 & \textbf{84.2} & \textbf{93.9} & 97.4 & \textbf{99.2} & \textbf{91.9} & \textbf{98.1} & \textbf{98.7} & \textbf{99.0} & \textbf{99.1} & \textbf{99.3} \\ \hline
    \end{tabular}
    \end{adjustbox}
    \label{table:sota2}
}

\vspace{1em}
% \end{adjustbox}
\caption{Performance comparisons with previous state-of-the-art approaches on image retrieval datasets.
For better comparison, values with the same color ({\color{Plum}purple}, {\color{blue}blue}, {\color{OliveGreen}green}, {\color{red} red}) have the same backbone, and embedding dimension (Dim), while bold text indicates the best performance within the same color.
$^\dagger$ denotes 256 input size for inference phase, while the rest use 224 input size.
$^\ddagger$ refers to non-conventional usage.}
\label{table:sota}
\end{center}
\vspace{-1em}
\end{table*}

Finally, we compare our proposed framework with the state-of-the-art approaches on four image retrieval datasets in Table ~\ref{table:sota1} and Table~\ref{table:sota2}.
To make a fair comparison, we put an experimental result using the same CNN backbone, input size, and output dimension with other approaches.
As each dataset has different properties, we choose the best performing configuration with ResNet-50 on each dataset by following the aforementioned rule in Section~\ref{sec:configurations_of_framework} and perform other experiments with the same configuration.
Even though BFE~\cite{dai2018batch} uses a 256 input size, and ours uses a 224 input size, the CGD framework gets higher performance on every dataset.
Overall, the CGD framework outperforms all the major benchmarks in the image retrieval tasks with a high margin.

%------------------------------------------------------------------------
\section{Conclusion}

In this paper, we have introduced a simple but powerful framework called CGD for image retrieval.
The CGD framework exploits multiple global descriptors to get an ensemble effect when it can be trained in an end-to-end manner.
Moreover, the proposed framework is flexible and expandable by global descriptors, CNN backbones, losses, and datasets.
We analyze the effectiveness of combined descriptor quantitatively and qualitatively.
Our extensive experiments show that exploiting multiple global descriptors lead to higher performance over the single global descriptor because combined descriptor can manipulate different types of feature properties.
Our framework performs the best on all the major image retrieval benchmarks considered.

%------------------------------------------------------------------------

{\small
\bibliographystyle{ieee}
\bibliography{egbib}
}

\end{CJK}
\end{document}